\documentclass{article}
\usepackage[nonatbib,final]{neurips_2022}
\usepackage[utf8]{inputenc} 
\usepackage[T1]{fontenc}    
\usepackage{hyperref}       
\usepackage{url}            
\usepackage{booktabs}       
\usepackage{amsfonts}       
\usepackage{nicefrac}       
\usepackage{microtype}      
\usepackage{graphicx}

\usepackage{subcaption}

\usepackage{multirow}

\usepackage[cmex10]{amsmath}
\usepackage{amssymb}

\usepackage{algorithmic}
\usepackage{array}
\usepackage{url}
\usepackage[noadjust]{cite}
\usepackage[table,xcdraw]{xcolor}
\usepackage{makecell}

\usepackage{authblk}
\begin{document}

\title{Cross-Geography Generalization of Machine Learning Methods for  Classification of Flooded Regions in Aerial Images}

\author[1]{\small Sushant Lenka}
\author[1]{\small Pratyush Kerhalkar}
\author[1]{\small Pranav Shetty}
\author[1]{\small Harsh Gupta}
\author[1]{\small Bhavam Vidyarthi}
\author[1, 2]{\small Ujjwal Verma}

\affil[1]{\footnotesize Department of Electronics and Communication Engg, Manipal Institute of Technology, Manipal Academy of Higher Education, India}
\affil[2]{\footnotesize Department of Electronics and Communication Engg, Manipal Institute of Technology Bengaluru, Manipal Academy of Higher Education, India}


\maketitle

\begin{abstract}
Identification of regions affected by floods is a crucial piece of information required for better planning and management of post-disaster relief and rescue efforts.  Traditionally, remote sensing images are analysed to identify the extent of damage caused by flooding. The data acquired from sensors onboard earth observation satellites are analyzed to detect the flooded regions, which can be affected by low spatial and temporal resolution. However, in recent years, the images acquired from Unmanned Aerial Vehicles (UAVs) have also been utilized to assess post-disaster damage. Indeed, a UAV based platform can be rapidly deployed with a customized flight plan and minimum dependence on the ground infrastructure. This work proposes two approaches for identifying flooded regions in UAV aerial images. The first approach utilizes texture-based unsupervised segmentation to detect flooded areas, while the second uses an artificial neural network on the texture features to classify images as flooded and non-flooded. Unlike  the existing works where the models are trained and tested on images of the same geographical regions, this work studies the performance of the proposed model in identifying flooded regions across geographical regions. An F1-score of 0.89 is obtained using the proposed segmentation-based approach which is higher than existing classifiers. The robustness of the proposed approach demonstrates that it can be utilized to identify flooded regions of any region with minimum or no user intervention.

\end{abstract}


\section{Introduction and Application Context}
With the increase in the frequency of occurrence of natural disasters, there is a need for better management and planning of post-disaster relief and rescue efforts. Identification of the region affected by the disaster is one of the crucial information required for planning post-disaster rescue efforts. The existing method for identifying disaster-affected areas depends on images acquired from sensors onboard Earth Observation Satellites\cite{ReviewFloodEO} \cite{FloodSAR}.  However, these images may be affected by cloud cover and have poor spatial resolution. Besides, the longer satellite visit time may adversely affect the post-disaster relief and rescue efforts. In recent years, images acquired from Unmanned Aerial Vehicles have been utilized for various applications such as environment monitoring, damage assessment etc \cite{UAVBuilding} \cite{Sharma2021476} \cite{Holla2020246}.

 



\par  For detecting the flooded areas, the aerial images are classified as flooded or non-flooded depending on the presence/absence of flooding in the image. In this paradigm, features (Color, textures etc) are first extracted from the images which are then fed to a classifier. In the last decade, several end-to-end Convolutional Neural Network (CNN) based approaches have been proposed, which eliminates the need for hand-engineering the features \cite{EmergencyNet}. In addition to the classification-based approach, a few studies use the segmentation results to categorize the image as flooded or non-flooded. In this approach, the images acquired from UAV are segmented and  assigned a semantic label (flooded/non-flooded) to each pixel in the image \cite{UAVFlood1} \cite{UAVFlood2} \cite{SushantIGARSS}. For instance, authors in \cite{UAVFlood1} used Fully Convolutional Networks (FCN) to segment the UAV images and detect the flooded regions. In separate work, in addition to segmenting the optical UAV images, the digital elevation model was analysed to detect the flooded region \cite{UAVFlood2}. In another study, an unsupervised segmentation algorithm was utilized to identify the flooded regions \cite{SushantIGARSS}. However, most existing classification and segmentation methods for identifying the flooded regions are limited to a particular region. While \cite{UAVFlood1, UAVFlood2} analysed images of flood-prone areas of North Carolina, USA, \cite{SushantIGARSS} analysed images acquired in Texas, USA, after Hurricane Harvey.  Extending these methods to new regions such as flood-prone regions of South Asia would require manually annotating additional images, which is a time-consuming process.  There is a need to develop methods for detecting flooded regions that can be generalized to unseen regions with minimum manual intervention.

The work proposes two texture features-based approaches to classify UAV aerial images as flooded or non-flooded across different geographical regions. Indeed, the texture features of pixels corresponding to water are different from that of other classes (Road, Buildings, greenery, etc), and, therefore can be used to identify flooded regions. In the first approach, an artificial neural network (ANN) is proposed in this work which distinguishes the images in texture feature space. In another approach, a texture feature-based unsupervised segmentation is proposed to first segment the images. The number of pixels corresponding to the flooded region is then used to classify the image as flooded or non-flooded. The segmentation-based approach is an extension of the method proposed in \cite{SushantIGARSS}. A texture-based unsupervised segmentation method was proposed in \cite{SushantIGARSS} to assess the severity of flooding in UAV aerial images of the \textit{same} geographical region. In this work, we adapt this method to ensure cross-geography generalization. The proposed method can be utilized to identify the flooded region and assess the flood severity in a UAV aerial image of \textit{any} region. In this work, the two proposed methods are compared with other existing machine learning classifiers and segmentation methods (UNet \cite{UNet}) to assess the performance of the proposed methods in detecting flooded regions \textit{across} different regions.

\section{Methodology}
\label{Sec:Methodology}

This work extends the texture-based unsupervised segmentation approach proposed in \cite{SushantIGARSS} for flooded region identification for cross-geography generalization. A summary of this unsupervised segmentation approach is presented in Section 2.1. In addition, this work proposes a neural network-based approach to identify flooded regions (Section 2.2 ).

\subsection{Texture Based Unsupervised Segmentation}
\label{SubSecKMeans}

The input image is first segmented into different regions using k-means segmentation. The segmentation is based on color (LAB color space) and texture information (Local Binary Pattern (LBP)). Including texture features ensure a more robust identification of regions covered with water. Indeed, pixels representing water have a distinctive texture appearance compared to other regions (roads, buildings, etc). The segmented region does not indicate which of the $k$ regions represent the flooded area. Therefore, each $k$ regions are compared with a reference flooded region using the histogram of Local Binary Pattern features. Subsequently, the input image is categorized as flooded or non-flooded based on the number of pixels corresponding to the flooded region. In \cite{SushantIGARSS}, the reference flooded the region and the input image were acquired in the same region. However, this work utilizes the reference flooded image of a completely different region. The use of the unsupervised segmentation approach along with reference images of a different region ensures that the proposed approach can be adapted to identify the flooded area of \textit{any} region with minimum user supervision.

\subsection{Artificial Neural Network}
\label{SubSec:ANN}
A customized Neural Network is proposed in this work for identifying the flooded regions in UAV aerial images. The input to the neural network is Local Binary Pattern (LBP) computed on the input image. Using LBP features instead of gray-scale or color intensities ensures that the network learns the texture pattern of the input image. The network consists of a sequence of seven dense layers for an input of 512 dimension LBP feature. The shape of $(i+1)^{th}$ dense layer is reduced by two from the previous $i^{th}$ layer. A dropout layer is also included after each dense layer. ReLu activation function was utilized for all the layers except the last layer. Sigmoid activation was used for the final layer. The penultimate layer had 8 nodes,  which were flattened and then fed to the final layer. The final layer had one neuron and sigmoid as the activation function for binary classification. If the activation score of the final layer was greater than 0.5, then the image was classified as flooded.  



\begin{figure}

     \centering
     \begin{subfigure}[b]{0.22\textwidth}
         \centering
         \includegraphics[width=\textwidth]{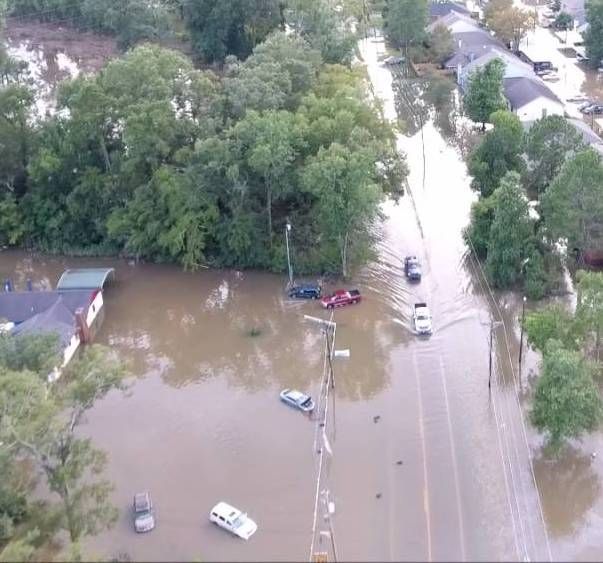}
         \label{fig:y equals x}
     \end{subfigure}
      \hfill
      \begin{subfigure}[b]{0.22\textwidth}
         \centering
         \includegraphics[width=\textwidth]{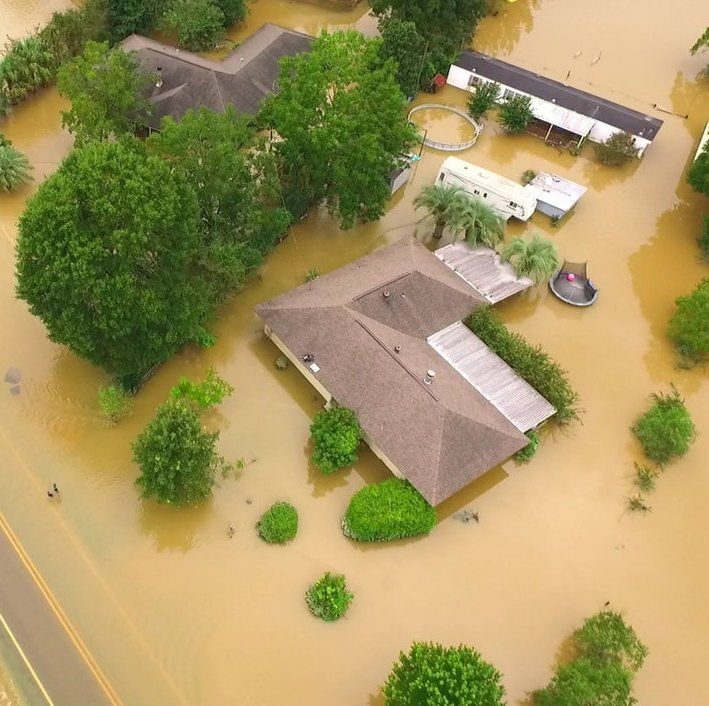} 
         \label{fig:five over x}
     \end{subfigure}
     \hfill
     \begin{subfigure}[b]{0.22\textwidth}
         \centering
         \includegraphics[width=\textwidth]{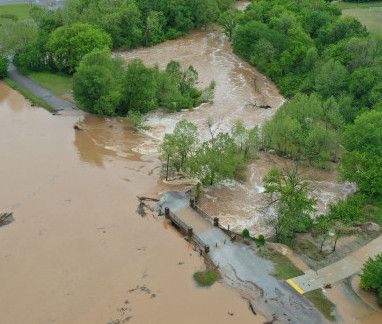} 
         \label{fig:five over x}
     \end{subfigure}
     \hfill
     \begin{subfigure}[b]{0.22\textwidth}
         \centering
         \includegraphics[width=\textwidth]{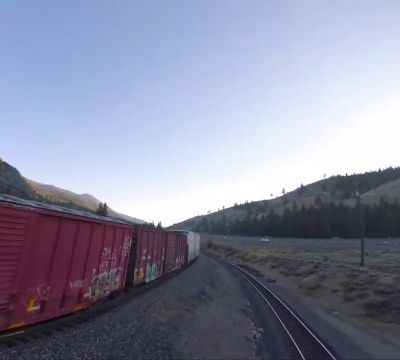} 
         \label{fig:five over x}
     \end{subfigure}
      
      
     \centering
     \begin{subfigure}[b]{0.23\textwidth}
         \centering
         \includegraphics[width=\textwidth]{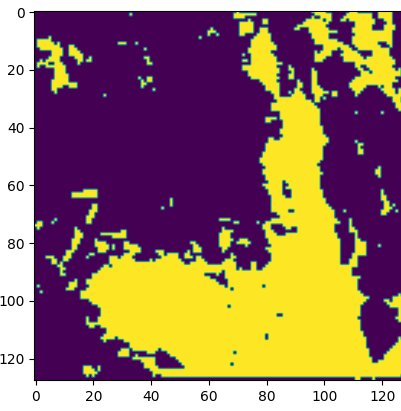}
         \label{fig:y equals x}
     \end{subfigure}
      \hfill
      \begin{subfigure}[b]{0.23\textwidth}
         \centering
         \includegraphics[width=\textwidth]{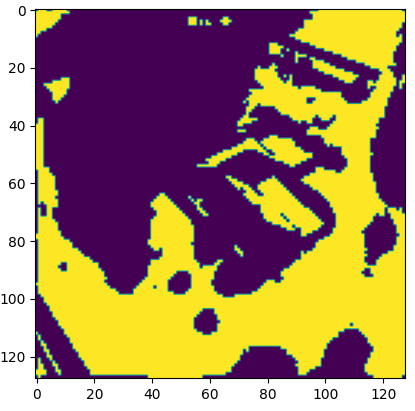} 
         \label{fig:five over x}
     \end{subfigure}
     \hfill
     \begin{subfigure}[b]{0.23\textwidth}
         \centering
         \includegraphics[width=\textwidth]{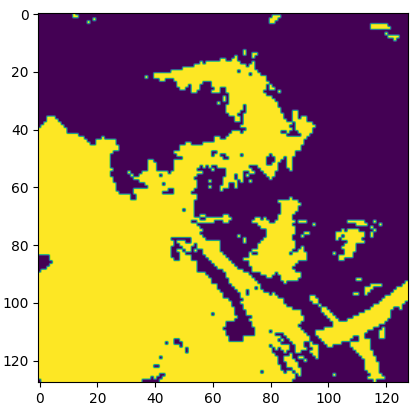} 
         \label{fig:five over x}
     \end{subfigure}
     \hfill
     \begin{subfigure}[b]{0.23\textwidth}
         \centering
         \includegraphics[width=\textwidth]{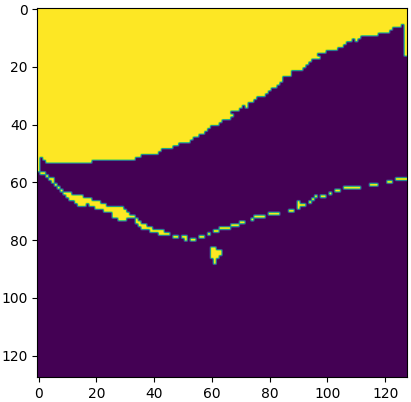} 
         \label{fig:five over x}
     \end{subfigure}

     
        \caption{Segmentation Result: The top row shows the input image while the bottom row shows the segmentation results obtained using the unsupervised segmentation approach. The yellow mask represents pixels corresponding to water. Note that the reference image in this case is from FloodNet dataset while the test images are from AIDER dataset.}
        \label{fig:three graphs}
\label{Fig:results}
 
\end{figure}

\section{Results}
\label{Sec:Results}

The proposed method is evaluated on a publicly available standard dataset viz. FloodNet \cite{Floodnet} and AIDER\cite{EmergencyNet,AIDER2}.  FloodNet \cite{Floodnet} contains images of the flooded areas acquired by UAVs after Hurricane Harvey. The dataset contains 3200 images and manual annotation for semantic segmentation of nine classes such as building-flooded, building-non-flooded, water, tree, etc. In this work, the image is considered flooded if the percentage of pixels corresponding to water is more than 25\%. The Aerial Image Dataset for Emergency Response applications (AIDER) dataset contains post-disaster UAV aerial images of four scenarios: Flood, Fire, Collapsed Buildings, and Traffic accidents. The AIDER dataset also contains images of normal cases (without any disaster). The images in this dataset were collected from various sources such as world-wide-web (e.g. google images, etc.), other aerial image datasets, and images collected using the authors' own UAV platform. This work considered images corresponding to only one disaster (flood). 

The two proposed methods were evaluated on FloodNet and AIDER datasets for classifying images as flooded or non-flooded. We estimated each classifier's precision, recall, F1-score, and accuracy to evaluate the classifiers' performance.   

The texture-based unsupervised segmentation approach requires estimating the optimal value of the segments ($k$). Table 1 shows the accuracy of the proposed segmentation approach for different values of $k$ with LBP and color features. Note that the reference images, in this case, are from the FloodNet dataset while the method is evaluated on images from the AIDER dataset. It can be observed that the highest accuracy is obtained with $k = 3$ for LAB colorspace. This work assumes that texture features can be utilized to distinguish between flooded and non-flooded images. To validate this hypothesis, the unsupervised segmentation approach was implemented with and without texture features (Table 2). Note that the reference images for the unsupervised segmentation approach are from the FloodNet dataset while the method is evaluated on images from the AIDER dataset. It can be seen that an F1-score of 0.89 was obtained with segmentation using texture features as compared to 0.69 for segmentation with only color features. Figure 1 shows the segmentation results obtained on the images of the AIDER dataset using FloodNet images as the reference. It can be seen that the proposed unsupervised segmentation approach can identify the region filled with water (first three columns) across different geographical regions. However, the approach incorrectly identifies the pixels corresponding to the sky as water in a few cases. 

In addition, a support vector machine (SVM) classifier was trained with the following feature vectors: 1) 128x128  histogram of LBP features, 2) the mean and variance of LBP histogram, and 3) color features only (Table 2). The SVM is trained and tested on the AIDER dataset. It can be seen that the inclusion of texture features results in a higher F1 score of the classifier. These results validate our hypothesis that texture features can aid in distinguishing flooded images from non-flooded images.

Table 3 shows the performance of SVM classifiers and the unsupervised segmentation approach evaluated on the same dataset. The performance of these classifiers is on expected lines with higher accuracy when texture features are considered. In addition, the proposed unsupervised segmentation-based approach is compared with a supervised segmentation-based approach viz UNet \cite{UNet}. Specifically, UNet is trained on images of the FloodNet dataset, and this trained model is used to segment images of AIDER dataset. Subsequently, the images are categorized as flooded or non-flooded based on the number of pixels corresponding to water in the segmented image. The poor F1-score (0.57) of this approach demonstrates that the segmentation method (UNet \cite{UNet}) fails to generalize across regions.

The performance of our proposed artificial neural network is evaluated for cross-geography generalization as follows: The network is trained on LBP features from images of a particular dataset and then evaluated on images of another dataset. The evaluation metric computed is shown in Table 4. It can be seen that an accuracy of 0.71 is obtained when trained on AIDER and tested on FloodNet, demonstrating cross-geography generalization. The network can also correctly classify when trained and tested on the same dataset. However, a lower accuracy is observed when trained on the FloodNet dataset and tested on AIDER. The reason is the limited number of images available in the FloodNet dataset compared to the AIDER dataset.

\begin{table}
\caption{Finding the optimum value of K and Colorspace for k-means. The reference image is from FloodNet dataset while accuracy is computed on images from the AIDER dataset. }
\centering
\begin{tabular}{|l|l|l|l|} 
\hline
\multicolumn{1}{|c|}{\textbf{Model}}                                                        & \multicolumn{1}{c|}{\textbf{Colorspace}} & \multicolumn{1}{c|}{\textbf{K}} & \multicolumn{1}{c|}{\textbf{Accuracy}}  \\ 
\hline
\multirow{4}{*}{\begin{tabular}[c]{@{}l@{}}K Means\\with LBP Texture Features\end{tabular}} & LAB                                      & 3                               & 0.8289                                  \\ 
\cline{2-4}
                                                                                            & RGB                                      & 3                               & 0.80418                                 \\ 
\cline{2-4}
                                                                                            & LAB                                      & 4                               & 0.7129                                  \\ 
\cline{2-4}
                                                                                            & RGB                                      & 4                               & 0.655                                   \\
\hline
\end{tabular}
\end{table}


\begin{table}[]
\caption{Evaluating the effectiveness of texture features: Comparison between k-means and SVM with and without texture features. For unsupervised segmentation, the reference image is from the FloodNet dataset while accuracy is computed on images from the AIDER dataset.}
\begin{tabular}{|c|c|c|c|c|c|}
\hline
\textbf{Model}                      & \textbf{Features}                             & \textbf{Accuracy}            & \textbf{Precision}           & \textbf{Recall}              & \textbf{F1Score}                                   \\ \hline
                                    & LBP                                           & 0.8289                       & 0.8326                       & 0.956                        & \multicolumn{1}{l|}{{\color[HTML]{333333} 0.8902}} \\ \cline{2-6} 
\multirow{0.5}{*}{k-means(K=3, LAB)} & No Texture Features                           & 0.56733                      & 0.8935                       & 0.5717                       & 0.6973                                             \\ \hline
                                    & \makecell{All 128x128 \\ LBP Histogram \\ Datapoints}         & \cellcolor[HTML]{FFFFFF}0.86 & \cellcolor[HTML]{FFFFFF}0.88 & \cellcolor[HTML]{FFFFFF}0.86 & \cellcolor[HTML]{FFFFFF}0.81                       \\ \cline{2-6} 
                                    & \makecell{Mean and Variance \\ of LBP Histogram \\ Datapoints} & 0.84                         & 0.7                          & 0.84                         & 0.76                                               \\ \cline{2-6} 
\multirow{1}{*}{SVM}               & Without Texture Features                      & 0.73                         & 0.81                         & 0.66                         & 0.7                                                \\ \hline
\end{tabular}
\end{table}

\begin{table}[]
\caption{Performance Evaluation of Classifiers (SVM), unsupervised segmentation (k-means) and supervised segmentation (UNet) based approaches.}
\begin{tabular}{|
>{\columncolor[HTML]{FFFFFF}}c |
>{\columncolor[HTML]{FFFFFF}}c |
>{\columncolor[HTML]{FFFFFF}}c |
>{\columncolor[HTML]{FFFFFF}}c |
>{\columncolor[HTML]{FFFFFF}}c |
>{\columncolor[HTML]{FFFFFF}}c |}
\hline
\textbf{Model}                                                                         & \textbf{Dataset} & \textbf{Accuracy} & \textbf{Precision} & \textbf{Recall} & \textbf{F1 Score} \\ \hline
\begin{tabular}[c]{@{}c@{}}SVM -\\ All 128x128 LBP\\ Histogram Datapoints\end{tabular} & AIDER            & 0.86              & 0.88               & 0.86            & 0.81              \\ \hline
\begin{tabular}[c]{@{}c@{}}K - Means\\ (K = 3, LBP and LAB Colorspace)\end{tabular}    & \textit{AIDER}   & 0.8289            & 0.8326             & 0.956           & 0.8902            \\ \hline
UNet Trained on FloodNet Dataset                                                       & AIDER            & 0.4528            & 0.673              & 0.507           & 0.578             \\ \hline
\begin{tabular}[c]{@{}c@{}}SVM with\\ texture features\end{tabular}                    & Floodnet         & 0.71              & 0.74               & 0.94            & 0.828             \\ \hline
\begin{tabular}[c]{@{}c@{}}SVM without\\ texture features\end{tabular}                 & Floodnet         & 0.65              & 0.51               & 0.71            & 0.6               \\ \hline
\end{tabular}
\end{table}

\begin{table}[]
\caption{Evaluating the performance of the Artificial Neural Network based approach in classifying images across different datasets.}
\centering
\begin{tabular}{|c|c|c|c|}
\hline
\textbf{Training Set} & \textbf{Testing Set} & \textbf{Accuracy} & \textbf{F1 Score} \\ \hline
AIDER                 & AIDER                & 0.8127            & 0.9125            \\ \hline
AIDER                 & Floodnet             & 0.7143            & 0.7034            \\ \hline
Floodnet              & Floodnet             & 0.6835            & 0.6138            \\ \hline
Floodnet              & AIDER                & 0.1905            & 0.2172            \\ \hline
\end{tabular}
\end{table}

\section{Conclusion}
This work proposes two texture features-based approaches for classifying UAV aerial images as flooded or non-flooded. It is observed that the texture of pixels representing water differs from that of other classes (buildings, roads, etc), which helps classify flooded images.  In the first approach, a texture-based unsupervised segmentation approach is proposed, where the input image is segmented with k-means segmentation using texture and color features. Subsequently, the segment corresponding to the flooded region is identified with the help of a few reference images. Finally, the number of pixels corresponding to water pixels is used to categorize images as flooded or non-flooded. In this work, we demonstrate that the reference image need not be acquired in the same geographical region as the input image. Besides, an artificial neural network approach is also presented, which categorizes the image in the texture feature space. The proposed method is evaluated on two standard datasets: FloodNet and AIDER. Our experimental setting focused on cross-geography generalization. Specifically, the results obtained demonstrate that the two proposed models trained on one dataset can be used to classify images of the other dataset. An F1-score of 0.89 was obtained using the proposed unsupervised segmentation approach when the reference image was from FloodNet while the test image was from the AIDER dataset. This result demonstrates the effectiveness of the proposed method of classifying flooded images across different regions.

 \bibliographystyle{plain}
\bibliography{Mendeley.bib}
\clearpage
 \appendix
 \section{Appendix}
 
 The hyper-parameters for the SVM model are determined using the standard grid search approach. The configuration used is shown in Table 5. We also evaluated the performance of multiple classifiers on classifying images as flooded or non-flooded. The accuracy and F1-score of these classifiers are shown in Table 6. The dataset utilized for training and testing these classifiers are also mentioned.  It may be noted that the highest F1 score of 0.85 is obtained with Passive Aggressive Classifiers trained on AIDER dataset and tested on FloodNet dataset.

\begin{table}[!h]

\centering
\caption{Parameter optimization using GridSearch for SVM Model.The final parameter configuration selected is shown in bold.}
\begin{tabular}{|l|l|}
\hline
\multicolumn{1}{|c|}{\textbf{Parameter}} & \multicolumn{1}{c|}{\textbf{Configuration}} \\ \hline
Gamma                                    & 1, \textbf{0.1}, 0.01, 0.001, 0.0001       \\ \hline
Penalty                                  & \textbf{0.1}, 1, 10, 100, 1000                       \\ \hline
Kernel                                   & linear, nonlinear, polynomial, \textbf{RBF},sigmoid  \\ \hline

\end{tabular}
\end{table}

\begin{table}[!h]
\caption{Performance evaluation of existing classifiers in classifying flooded images. The dataset utilized for training and testing these classifiers are also mentioned.}
\begin{tabular}{|c|c|c|c|c|}
\hline
\textbf{Model}                & \textbf{Accuracy} & \textbf{F1 Score} & \textbf{Training   Dataset} & \textbf{Testing Dataset} \\ \hline
DecisionTree                  & 0.77              & 0.78              & AIDER                       & AIDER                    \\ \hline
SGDClassifier                 & 0.8               & 0.81              & AIDER                       & AIDER                    \\ \hline
AdaBoostClassifier            & 0.86              & 0.85              & AIDER                       & AIDER                    \\ \hline
PassiveAggressiveClassifier   & 0.67              & 0.71              & AIDER                       & AIDER                    \\ \hline
LGBMClassifier                & 0.88              & 0.85              & AIDER                       & AIDER                    \\ \hline
BaggingClassifier             & 0.86              & 0.84              & AIDER                       & AIDER                    \\ \hline
BernoulliNB                   & 0.86              & 0.84              & AIDER                       & AIDER                    \\ \hline
Perceptron                    & 0.66              & 0.7               & AIDER                       & AIDER                    \\ \hline
NearestCentroid               & 0.85              & 0.83              & AIDER                       & AIDER                    \\ \hline
XGBClassifier                 & 0.86              & 0.83              & AIDER                       & AIDER                    \\ \hline
LinearDiscriminantAnalysis    & 0.86              & 0.81              & AIDER                       & AIDER                    \\ \hline
RidgeClassifier               & 0.86              & 0.81              & AIDER                       & AIDER                    \\ \hline
GaussianNB                    & 0.84              & 0.79              & AIDER                       & AIDER                    \\ \hline
ExtraTreesClassifier          & 0.79              & 0.75              & Floodnet                    & Floodnet                 \\ \hline
PassiveAggressiveClassifier   & 0.64              & 0.66              & Floodnet                    & Floodnet                 \\ \hline
SGDClassifier                 & 0.62              & 0.64              & Floodnet                    & Floodnet                 \\ \hline
Perceptron                    & 0.62              & 0.64              & Floodnet                    & Floodnet                 \\ \hline
NearestCentroid               & 0.76              & 0.69              & Floodnet                    & Floodnet                 \\ \hline
RidgeClassifier               & 0.74              & 0.65              & Floodnet                    & Floodnet                 \\ \hline
LGBMClassifier                & 0.74              & 0.65              & Floodnet                    & Floodnet                 \\ \hline
GaussianNB                    & 0.74              & 0.65              & Floodnet                    & Floodnet                 \\ \hline
RandomForestClassifier        & 0.71              & 0.6               & Floodnet                    & Floodnet                 \\ \hline
BaggingClassifier             & 0.69              & 0.58              & Floodnet                    & Floodnet                 \\ \hline
AdaBoostClassifier            & 0.67              & 0.57              & Floodnet                    & Floodnet                 \\ \hline
CalibratedClassifierCV        & 0.77              & 0.73              & Floodnet                    & AIDER                    \\ \hline
QuadraticDiscriminantAnalysis & 0.6               & 0.62              & Floodnet                    & AIDER                    \\ \hline
DecisionTreeClassifier        & 0.62              & 0.64              & AIDER                       & Floodnet                 \\ \hline
PassiveAggressiveClassifier   & 0.74              & 0.85              & AIDER                       & Floodnet                 \\ \hline
LinearSVC                     & 0.64              & 0.78              & AIDER                       & Floodnet                 \\ \hline
LabelSpreading                & 0.71              & 0.6               & AIDER                       & Floodnet                 \\ \hline
\end{tabular}
\end{table}

\end{document}